\documentclass[10pt,twocolumn,letterpaper]{article}

\usepackage{iccv}
\usepackage{times}
\usepackage{epsfig}
\usepackage{graphicx}
\usepackage{amsmath}
\usepackage{amssymb}

\usepackage{booktabs}
\usepackage{verbatim}
\usepackage{subcaption}
\usepackage{multirow}
\usepackage{bm}

\makeatletter
\@namedef{ver@everyshi.sty}{}
\makeatother
\usepackage{pgfplots}
\usepackage{pgfplots}\pgfplotsset{compat=1.9}

\newcommand*{\affaddr}[1]{#1} 
\newcommand*{\affmark}[1][*]{\textsuperscript{#1}}
\newcommand*{\email}[1]{#1}

\newcounter{alphasect}
\def\alphainsection{0}

\let\oldsection=\section
\def\section{%
  \ifnum\alphainsection=1%
    \addtocounter{alphasect}{1}
  \fi%
\oldsection}%

\renewcommand\thesection{%
 \ifnum\alphainsection=1%
   \Alph{alphasect}%
 \else
   \arabic{section}%
 \fi%
}%

\newenvironment{alphasection}{%
  \ifnum\alphainsection=1%
    \errhelp={Let other blocks end at the beginning of the next block.}
    \errmessage{Nested Alpha section not allowed}
  \fi%
  \setcounter{alphasect}{0}
  \def\alphainsection{1}
}{%
  \setcounter{alphasect}{0}
  \def\alphainsection{0}
}%

\usepackage[breaklinks=true,bookmarks=false]{hyperref}

\iccvfinalcopy 

\def\iccvPaperID{****} 

\ificcvfinal\fi

\begin{document}

\title{Synthesizing Realistic Image Restoration Training Pairs: A Diffusion Approach}


\author{%
Tao Yang\affmark[1], Peiran Ren\affmark[1], Xuansong Xie\affmark[1], and Lei Zhang\affmark[2]\footnotemark\\
\affaddr{\affmark[1]DAMO Academy, Alibaba Group}\\
\affaddr{\affmark[2]Department of Computing, The Hong Kong Polytechnic University}\\
\email{\tt\small{yangtao9009@gmail.com, peiran\_r@sohu.com, xingtong.xxs@taobao.com, cslzhang@comp.polyu.edu.hk}}\\
}

\maketitle
\ificcvfinal\thispagestyle{empty}\fi

\begin{abstract}
   In supervised image restoration tasks, one key issue is how to obtain the aligned high-quality (HQ) and low-quality (LQ) training image pairs. Unfortunately, such HQ-LQ training pairs are hard to capture in practice, and hard to synthesize due to the complex unknown degradation in the wild. While several sophisticated degradation models have been manually designed to synthesize LQ images from their HQ counterparts, the distribution gap between the synthesized and real-world LQ images remains large. We propose a new approach to synthesizing realistic image restoration training pairs using the emerging denoising diffusion probabilistic model (DDPM). 
   First, we train a DDPM, which could convert a noisy input into the desired LQ image, with a large amount of collected LQ images, which define the target data distribution. Then, for a given HQ image, we synthesize an initial LQ image by using an off-the-shelf degradation model, and iteratively add proper Gaussian noises to it. Finally, we denoise the noisy LQ image using the pre-trained DDPM to obtain the final LQ image, which falls into the target distribution of real-world LQ images. Thanks to the strong capability of DDPM in distribution approximation, the synthesized HQ-LQ image pairs can be used to train robust models for real-world image restoration tasks, such as blind face image restoration and blind image super-resolution. Experiments demonstrated the superiority of our proposed approach to existing degradation models. Code and data will be released.
\end{abstract}

\section{Introduction}
\label{sec:intro}
Deep neural networks (DNNs) \cite{lecun2015deep} have been successfully used in a variety of computer vision tasks, including image classification \cite{he2016resnet}, object detection \cite{ren2015fasterrcnn}, segmentation \cite{he2017maskrcnn}, as well as image restoration \cite{dong2014srcnn}. In supervised learning, the amount and quality of labeled training data will largely affect the practical performance of trained DNN models. This problem becomes more crucial in real-world image restoration tasks \cite{yang2021gpen,zhang2021bsrgan}, where the aligned high-quality (HQ) and low-quality (LQ) training image pairs are very difficult to capture in practice. 
Early works mainly resort to using bicubic downsampling or some simple degradation models \cite{lai2017lapsrn} to synthesize LQ images from their HQ counterparts, which however can only cover a very small set of degradation types. The real-world LQ images can suffer from many factors, including but not limited to low resolution, blur, noise, compression, \etc., which are too complex to be explicitly modeled. As a result, the DNN models trained on synthesized HQ-LQ training pairs can hardly perform well on real-world LQ images. 

To alleviate the above difficulties, some works have been proposed to first predict the degradation parameters \cite{gu2019ikc,guo2019cbdnet,jiang2021fbcnn} and then restore the HQ image with them in a non-blind way. They work well on some specific non-blind deblurring  \cite{gu2019ikc}, denoising \cite{guo2019cbdnet} and JPEG artifacts removal \cite{jiang2021fbcnn} tasks. However, the degradation of real-world LQ images are often unknown and cannot be pre-defined, making it is hard, if not possible, to predict accurate degradation parameters. Some researchers attempted to collect real-world LQ-HQ pairs \cite{cai2019realsr,wei2020cdc} by using long-short camera focal lengths, which only work in applications where similar photographing devices are used. Recently, a couple of handcrafted degradation models \cite{zhang2021bsrgan,wang2021realesrgan} have been proposed to model complex real-world degradations. Zhang \etal \cite{zhang2021bsrgan} randomly shuffled blur, downsampling and noise degradations to form a complex combination. Wang \etal \cite{wang2021realesrgan} developed a high-order degradation model. While these methods can simulate a much larger scope of degradation types and have shown impressive progress in handling LQ images in the wild, the distribution gap between the synthesized and real-world LQ images remains large \cite{li2022face2natual}. 

With the rapid advancement of deep generative networks \cite{goodfellow2014gan,karras2019stylegan2,ho2020ddpm}, methods have been developed to learn how to synthesize LQ images from their HQ counterparts. Lugmayr \etal \cite{lugmayr2019unsupervised} learned a domain distribution network using unpaired data and then built HQ-LQ pairs with it. Similarly, Fritsche \etal \cite{fritsche2019frequency} synthesized LQ images by using DSGAN to introduce natural image characteristics. Luo \etal \cite{luo2022pdmsr} proposed a probabilistic degradation model (PDM) to describe different degradations. Most recently, Li \etal \cite{li2022face2natual} developed the ReDegNet to model real-world degradations using paired face images and transfer them to produce LQ natural images. However, this method largely relies on the HQ faces generated by blind face restoration models \cite{yang2021gpen,wang2021gfpgan}, which limits its applications.

In this work, we revisit the problem of HQ-LQ image pair synthesis, which is critical to the many image restoration tasks such as blind face restoration and real-world image super-resolution. Our idea is to seamlessly integrate the advantages of handcrafted degradation models and deep generative networks. We first train a generator, \ie, the emerging denoising diffusion probabilistic model (DDPM) \cite{ho2020ddpm}, by using a large amount of real-world LQ images collected from the Internet. The trained DDPM can be used to generate realistic LQ images. When building HQ-LQ image pairs, we first adopt a handcrafted degradation model to synthesize initial LQ images from the input HQ images, which may fall into the distribution of real-world LQ images. To reduce the distribution gap, the initially synthesized LQ images are added with Gaussian noise and then denoised by the pre-trained DDPM. After a few steps, the distribution of synthesized LQ images will become closer and closer to the distribution of real-world LQ images. Finally, a set of realistic HQ-LQ training pairs can be synthesized, and they can be used to train robust DNN models for image restoration tasks. 

In summary, in this work we present a novel diffusion approach to synthesizing realistic training pairs for image restoration, targeting at shortening the distribution gap between synthetic and real-world LQ data. The synthesized HQ-LQ training pairs can be used for various downstream real-world image restoration tasks, as validated in our experiments of blind image restoration and blind image super-resolution. To the best of our knowledge, this is the first image degradation modeling method based on diffusion models. Codes will be made publicly available.

\section{Related Work}
\subsection{Degradation Modeling}
Image degradation modeling is of great importance for many downstream tasks such as blind face restoration (BFR) \cite{yang2021gpen} and blind image super-resolution (BISR) \cite{zhang2021bsrgan}. 
Bicubic downsampling has been popularly adopted in the research of single image super-resolution (SISR) \cite{dong2014srcnn,ledig2017srgan,wang2018esrgan}. Though it provides a common platform for comparison of SISR algorithms, models trained with this degradation strategy are of little avail, especially for real-world applications. Some works turned to use the classical degradation models \cite{lai2017lapsrn}, which take the commonly observed noise, blur, and downsampling degradations into consideration. Unfortunately, they are far from enough to describe the complex unknown degradations in real-world LQ images. 

Zhang \etal \cite{zhang2021bsrgan} proposed a random shuffling strategy to construct more complex degradations of blur, downsampling and noise. Concurrently, Wang \etal \cite{wang2021realesrgan} designed a high-order degradation model with several repeated degradation processes using the classical degradation model. These two methods managed to simulate diverse degradation combinations. 
Considering that real-world degradations are too complex to be explicitly modeled, some researchers attempted to learn a network to implicitly approximate the degradation process \cite{lugmayr2019unsupervised,fritsche2019frequency,maeda2020pseudo,wei2021dasr,wang2021unsupervised,wang2021unsupervisedsr,luo2022pdmsr,li2022face2natual}. Nonetheless, it is challenging to learn such models due to the high illness of the problem. 

\textbf{Blind Face Restoration.}
Face image restoration has attracted a lot of attentions \cite{li2018gfrnet,hu20203dprior,yang2021gpen,wang2021gfpgan,zhou2022codeformer}. Since facial images have specific structures, it is feasible to restore a clear face image from real-world degraded observations \cite{yang2021gpen,wang2021gfpgan,zhou2022codeformer}. Many BFR methods have been proposed by resorting to exemplar images \cite{li2018gfrnet}, 3D facial priors \cite{hu20203dprior}, and facial component dictionaries \cite{li2020dfdnet}. Recently, generative face prior has been widely used and shown powerful capability in BFR tasks \cite{yang2021gpen,wang2021gfpgan}. It has been demonstrated that many LQ face images in the wild can be robustly restored. Researchers have also found that a learned discrete codebook prior can better reduce the uncertainty and ambiguity of face restoration mapping \cite{zhou2022codeformer}. 

\textbf{Blind Image Super-resolution.}
While significant progress has been achieved for non-blind SISR \cite{dong2014srcnn,ledig2017srgan,wang2018esrgan}, blind image super-resolution (BISR) remains very difficult. Bell-Kligler \etal \cite{bell-kligler2019kernelgan} introduced KernelGAN to estimate the blur kernel and then restore images based on it, which is prone to errors of estimated kernels. Following works jointly predicted the blur kernel and the HQ image. Gu \etal \cite{gu2019ikc} suggested an iterative correction scheme to achieve better results. 
Those methods focused on blur kernel estimation, while images in the wild suffer from much more complex degradations other than blur. Cai \etal \cite{cai2019realsr} and Wei \emph{et al.} \cite{wei2020cdc} respectively established an SISR dataset with paired LQ-HQ images collected by zooming camera lens. However, the model can be hardly generalized to other photographing devices. Some works exploited DNNs to learn the degradation process with unpaired data \cite{lugmayr2019unsupervised,fritsche2019frequency,maeda2020pseudo,wei2021dasr,wang2021unsupervised,wang2021unsupervisedsr,luo2022pdmsr,li2022face2natual}. Lugmayr \emph{et al.} \cite{lugmayr2019unsupervised} employed a cycle consistency loss to learn a distribution mapping network. Fritsche \emph{et al.} \cite{fritsche2019frequency} proposed DSGAN to generate LQ images. Luo \etal \cite{luo2022pdmsr} modeled the degradation as a random variable and learned its distribution. Li \emph{et al.} \cite{li2022face2natual} transferred the real-world degradations learned from face images to natural images. Although those methods have shown impressive results in some cases, their overall generalization performance in the wild remains limited. 

\subsection{Deep Generative Network}
The generative adversarial networks (GANs) \cite{goodfellow2014gan} have demonstrated much more powerful capability to synthesize HQ images than likelihood-based methods such as variational autoencoders (VAE) \cite{kingma2014vae}, autoregressive models \cite{oord2016prnn} and flows \cite{rezende2015flow}. 
Recently, diffusion probabilistic models (DPMs) have emerged and shown promising performance in tasks of image generation \cite{ho2020ddpm}, image inpainting \cite{rombach2022latent}, image-to-image translation \cite{saharia2022palette}, text-to-image generation \cite{ramesh2022dalle2}, text-to-video generation \cite{ho2022imagenvideo}, etc. In particular, high quality image synthesis results were presented in \cite{ho2020ddpm}, which was extended and improved in \cite{nichol2021improved,dhariwal2021guided}. Despite the great success, DPMs require hundreds of steps to simulate a Markov chain. Song \emph{et al.} \cite{song2021ddim} proposed a denoising diffusion implicit model (DDIM) to accelerate the sampling speed. Very recently, DPMs have shown impressive results in text-to-image/video generation \cite{ramesh2022dalle2,saharia2022imagen,ho2022imagenvideo}. 

\begin{figure}[t!]
\centering
\includegraphics[width=0.48\textwidth]{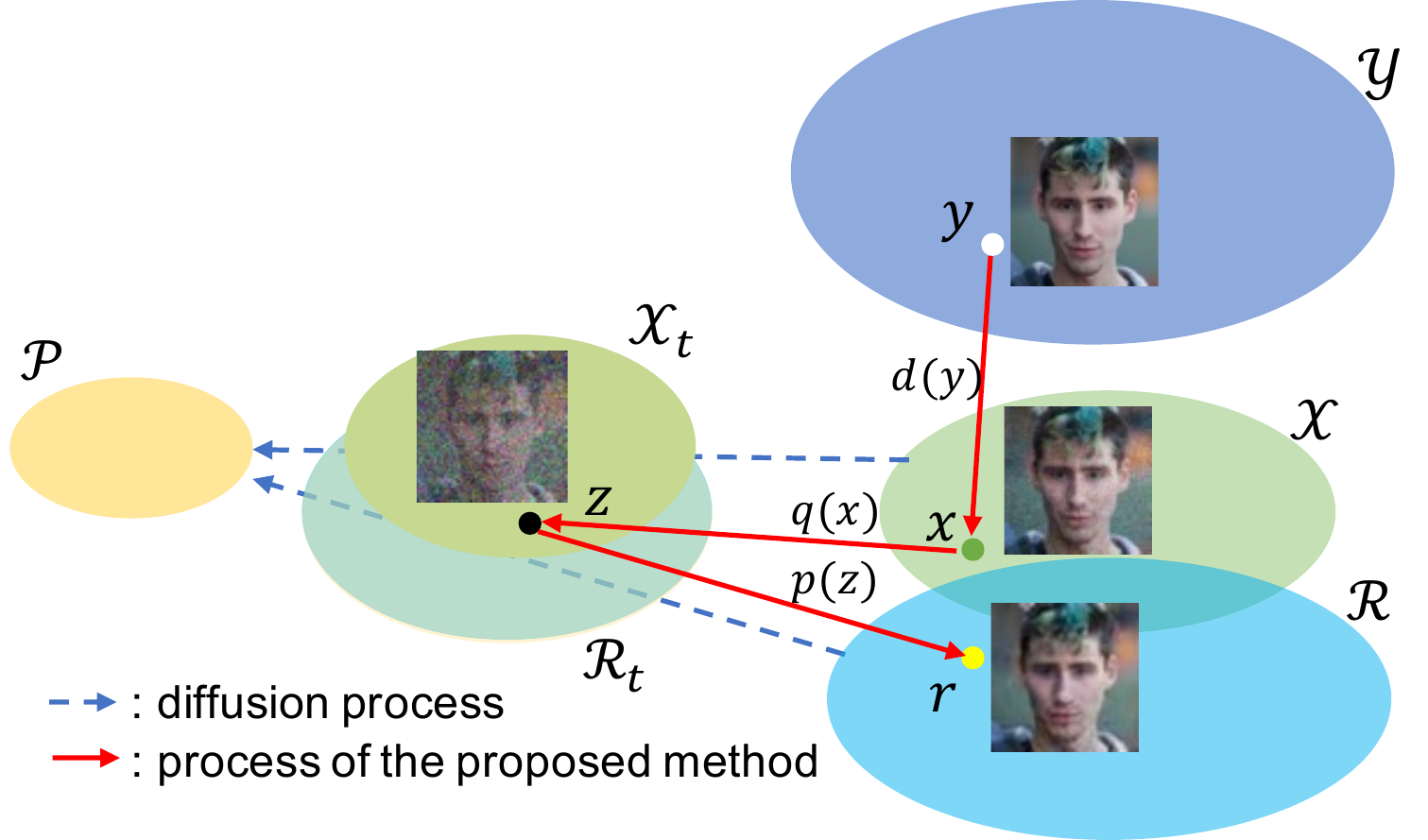}
\caption{Illustration of the proposed HQ-LQ image pair synthesis process. Please refer to Sec. \ref{sec:motivation} for details.}
\label{fig:motivation}
\vspace*{-3mm}
\end{figure}

\section{Proposed Method}
\subsection{Problem Formulation and Framework}
\label{sec:motivation}
Let's denote by $\mathcal{Y}$ the space of original HQ images, by $\mathcal{X}$ the space of synthetic LQ images, and by $\mathcal{R}$ the space of real-world LQ images. Generally speaking, the spaces $\mathcal{X}$ and $\mathcal{R}$ are partially overlapped because  synthetic LQ images can have similar statistics to the real-world LQ images. How large the overlapped subspace between $\mathcal{X}$ and $\mathcal{R}$ will be depends on the employed degradation model, denoted by $d$. For example, Zhang \emph{et al.} \cite{zhang2021bsrgan} and Wang \emph{et al.} \cite{wang2021realesrgan} handcrafted sophisticated models to synthesize LQ data from their corresponding HQ images. Some works resorted to learning a mapping network from $\mathcal{Y}$ to $\mathcal{R}$ by unsupervised learning \cite{lugmayr2019unsupervised,fritsche2019frequency,maeda2020pseudo,wei2021dasr,luo2022pdmsr} or transferring degradations \cite{li2022face2natual}. Nonetheless, the distributions gap between synthesized and real-world LQ images remains large. 

Different from these previous efforts \cite{lugmayr2019unsupervised,wei2021dasr,luo2022pdmsr,zhang2021bsrgan,wang2021realesrgan,li2022face2natual}, we aim to integrate the advantages of handcrafted degradation models and deep generative networks, more specifically, the denoising diffusion probabilistic models (DDPM) \cite{ho2020ddpm}, to synthesize realistic training pairs. As shown in Fig.~\ref{fig:motivation}, given an HQ image $y\in\mathcal{Y}$, we first adopt a state-of-the-art degradation model \cite{zhang2021bsrgan,wang2021realesrgan} $d$ to generate an initial LQ image $x\in\mathcal{X}$, \ie, $x=d(y)$. Usually, $x$ shares some similar statistics and appearances to images in $\mathcal{R}$, while it still has certain distance to space $\mathcal{R}$. To make $x$ closer to $\mathcal{R}$, we train a DDPM, which defines a Markov chain of diffusion steps to add random noise to data and then reverses the diffusion process to construct desired data samples from noise, to further transform $x$ into $r\in\mathcal{R}$. 

The diffusion process will gradually convert the data of a complex distribution into the data of a simpler prior distribution $\mathcal{P}$, \eg, the isotropic Gaussian distribution. In other words, the distributions of images in $\mathcal{X}$ and $\mathcal{R}$ would become closer and closer during the diffusion process, and approach to the same isotropic Gaussian distribution finally. As depicted in Fig.~\ref{fig:motivation}, we denote by $\mathcal{X}_t$ and $\mathcal{R}_t$ respectively the latent distributions of $\mathcal{X}$ and $\mathcal{R}$ after diffusing $t$ steps. One can see that the domain gap between $\mathcal{X}_t$ and $\mathcal{R}_t$ becomes more and more indistinguishable. 

Based on the above consideration, we first diffuse $x$ by $t$ steps to generate $z$, \ie, $z=q(x)$, which is more likely to fall into space $\mathcal{R}_t$. We then reverse $z$ to obtain the final LQ image $r$, \ie, $r=p(z)$, using DDPM pre-trained on real-world LQ images. The whole HQ to LQ image synthesis process is illustrated by the solid red line in Fig.~\ref{fig:motivation}. Due to the powerful distribution approximation capability of DDPM, it is anticipated that the final LQ image $r$ will fall into the space $\mathcal{R}$ of real-world LQ images. Consequently, aligned HQ-LQ image pairs, \eg, $(y,r)$, can be obtained and used to train more robust image restoration models than those trained with $(y,x)$. 

\begin{figure}[t!]
\centering
\includegraphics[width=0.48\textwidth]{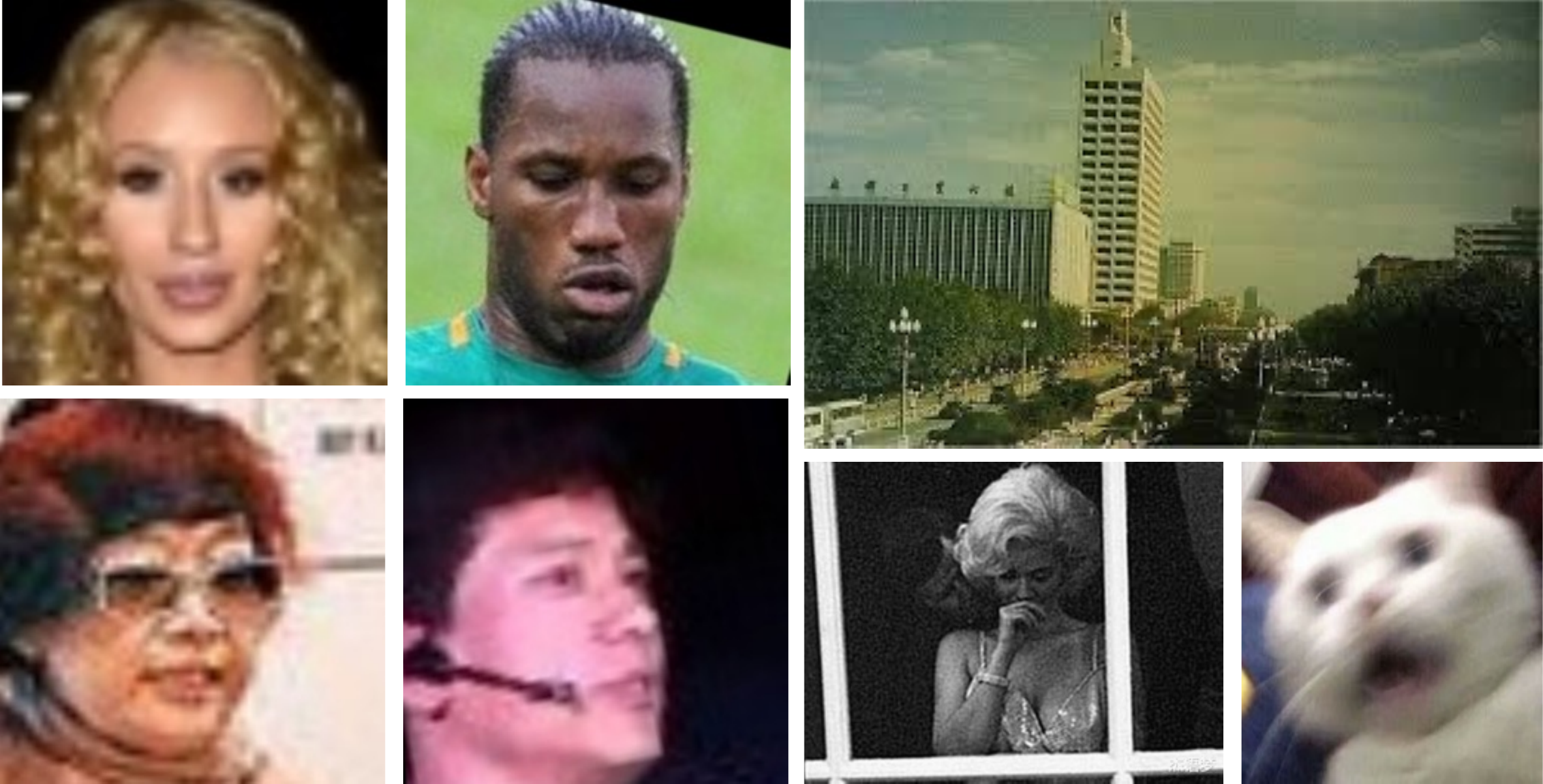}
\vspace*{-3mm}
\caption{Sample face and natural scene images in our collected degraded image dataset (DID).}
\label{fig:did}
\vspace*{-3mm}
\end{figure}

\begin{figure*}[t!]
\centering
\includegraphics[width=0.9\textwidth]{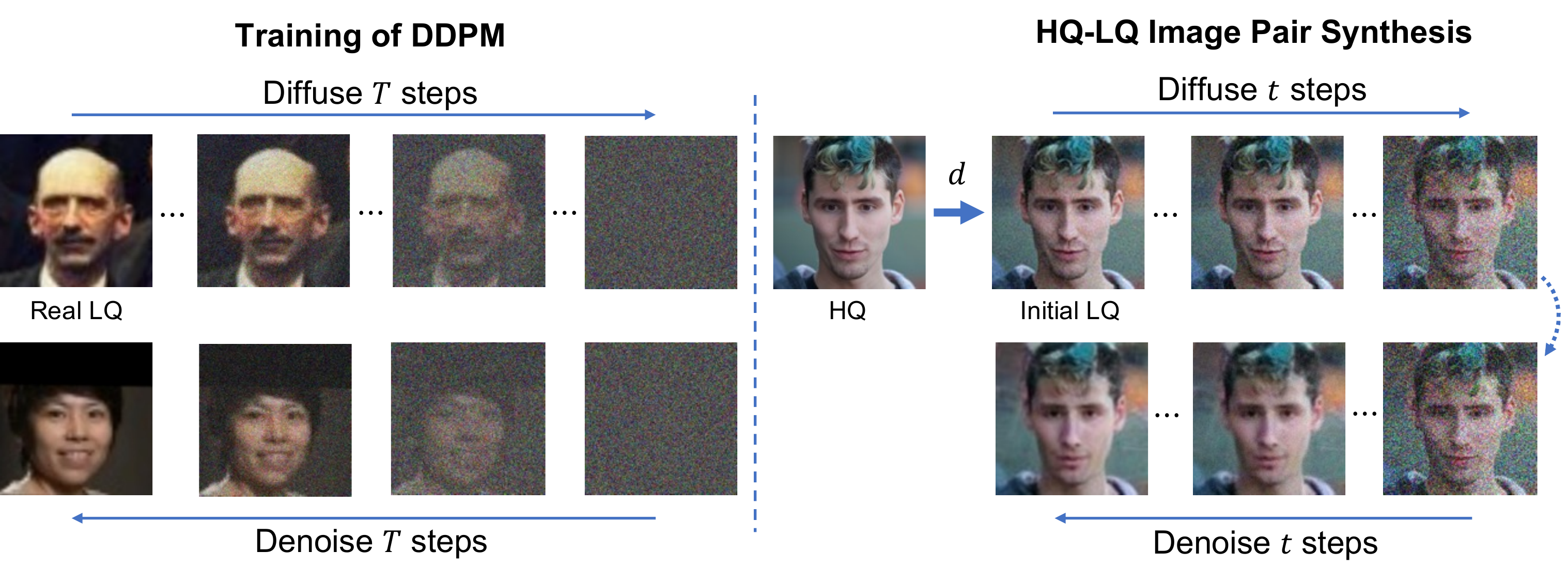}
\vspace{-3mm}
\caption{Left: illustration of the training process of DDPM (refer to Sec.~\ref{sec:training} for details). Right: illustration of the HQ-LQ image pair generation process by using the pre-trained DDPM (refer to Sec.~\ref{sec:synthesis} for details).}
\label{fig:arch}
\vspace*{-3mm}
\end{figure*}

\subsection{DDPM Model Training}
\label{sec:training}
\textbf{Degraded Image Dataset.} Our method employs a DDPM to convert the HQ image into the LQ one, which is expected to meet the distribution of real-world LQ images. Therefore, a large-scale real-world LQ image dataset is needed to train the DDPM model first. Since most of the publicly available datasets are composed of HQ images \cite{timofte2017sr,karras2019stylegan2} or synthetic LQ ones \cite{zhang2018lpips}, we build a large-scale degraded image dataset (DID) from scratch, which consists of $70,000$ degraded face images and $30,000$ degraded natural images. Some example images are shown in Fig.~\ref{fig:did}. 

We first collected a large number of images from the Internet, where each image contains at least one face. Then the facial portions are automatically detected, cropped and aligned \cite{deng2020retinaface}. Every face image is resized to have a resolution of $512\times 512$. We carefully prune the data to exclude occasional statues, paintings, \etc. In addition, we tend to not harvest too many faces in a single group of photos in order to diversify the degradation types. Finally, $70,000$ real-world degraded faces are collected. 

To tackle more general image restoration problems such as blind image super-resolution, we crawled another set of $30,000$ real-world natural images with diverse sizes and contents from the Internet. Specifically, we use more than $200$ keywords, which cover a large scope of categories, including human, animal, landscape, indoor scene, \etc., to search via Google Images. Different from face images, we keep the crawled images unchanged. 

\textbf{Denoising Diffusion Probabilistic Models.}
Given a data point from a real data distribution $x_0\sim q(x)$, it is interesting to learn a model distribution $p(x)$ that could approximate $q(x)$. To achieve this goal, DDPM \cite{ho2020ddpm} defines a forward diffusion process $q$, which produces a sequence of latents $\{x_1,x_2,...,x_T\}$ by adding Gaussian noise in $T$ steps. The step sizes are controlled by a variance schedule $\beta_t\in(0,1)$, where $t\in\{1,2,...,T\}$. There is:
\begin{align}
q(x_t|x_{t-1})&:=\mathbf{\mathcal{N}}(x_t:\sqrt{1-\beta_tx_{t-1}},\beta_t\mathbf{I}).
\end{align}
In particular, the above equation allows us to sample a latent at an arbitrary step $t$ directly, conditioned on the input $x_0$. Let $\alpha_t=1-\beta_t$ and $\bar{\alpha}_t=\prod_{i=1}^t\alpha_i$, we can write the marginal distribution as follows:
\begin{equation}
q(x_t|x_0)=\mathbf{\mathcal{N}}(x_t;\sqrt{\bar{\alpha}_t}x_0,(1-\bar{\alpha}_t)\mathbf{I}).
\end{equation}
With the help of reparameterization techniques \cite{kingma2014vae}, $x_t$ can be formulated as follows:
\begin{equation}
x_t=\sqrt{\bar{\alpha}_t}x_0+\sqrt{1-\bar{\alpha}_t}\epsilon, ~where~\epsilon\sim \mathcal{N}(\mathbf{0}, \mathbf{I}).
\label{eqn:x_02x_t}
\end{equation}
Since $\alpha_t\in(0,1)$, $x_T$ is equivalent to an isotropic Gaussian distribution when $T\to\infty$. Usually, $T$ is set to $1,000$. 

If the exact reverse distribution $q(x_{t-1}|x_t)$ is known, we will be able to recreate the original samples from a Gaussian noise input. Unfortunately, $q(x_{t-1}|x_t)$ depends on the entire data distribution. We need to learn a model $p_\theta$ to approximate it. It is worth noting that if $\beta_t$ is small enough, $q(x_{t-1}|x_t)$ will also be Gaussian, \ie,
\begin{align}
p(x_{t-1}|x_t)&:=\mathbf{\mathcal{N}}(x_{t-1};\mathbf{\mu}_\theta(x_t,t),\mathbf{\Sigma}_\theta(x_t,t)).
\end{align}


\textbf{Model Training.}
With the collected DID, we train a DDPM to approximate its distribution. As illustrated on the left side of Fig.~\ref{fig:arch}, during each optimization iteration, we diffuse a randomly sampled LQ image $x_0$ from DID for $t$ steps, where $t$ is randomly chosen from $\{0,1,,...,T-1\}$. According to Eqn.~\ref{eqn:x_02x_t}, we can easily compute $x_t$ with a randomly generated Gaussian noise $\epsilon$. In order to reverse the diffusing process, DDPM is designed to predict the noise $\epsilon$ on top of $x_t$ and $t$. To this end, we can parameterize $\epsilon_\theta(x_t,t)$ with a simplified loss function:
\begin{equation}
L_{simple}=E_{t,x_0,\epsilon}[||\epsilon-\epsilon_\theta(x_t,t)||].
\end{equation}
Different from Ho \etal \cite{ho2020ddpm}, we use $L_1$ instead of $L_2$ loss here because $L_1$ loss is usually more resistant to overfitting and encourages less predictive features. Particularly, for the task of BISR, we randomly crop patches of resolution $256\times 256$ from the original image as input $x_0$.

\subsection{HQ-LQ Image Pair Synthesis}
\label{sec:synthesis}
With the DDPM model trained in Sec. \ref{sec:training}, we are able to synthesize realistic LQ images from the HQ ones, as illustrated on the right side of Fig.~\ref{fig:arch}. We first apply a handcrafted degradation model to the input HQ image to obtain an initial LQ image, and then iteratively add proper Gaussian noises to the initially synthesized LQ image for $t$ steps, obtaining a noisy LQ image $x_t$. Note that $x_t$ can be directly calculated according to Eqn.~\ref{eqn:x_02x_t} without time-consuming iterations. This is because mathematically, the merge of two Gaussian noises will result in another Gaussian noise. 

We then employ the pre-trained DDPM to predict the Gaussian noise $\epsilon$ added to $x_t$. With the predicted noise, we obtain a predicted $x_0$, which incorporates $x_t$ to produce $x_{t-1}$. We iteratively perform the above operations for $t$ times and finally obtain a synthesized LQ image. Generally speaking, the quality of the output LQ image depends on the setting of $t$. Since the DDPM model is trained to generate samples that meet the target distribution of real-world LQ images, our synthesized LQ images are more realistic than the manually synthesized ones, as we discussed in Sec.~\ref{sec:motivation}. 

In the aforementioned synthesizing process, one can see that the handcrafted degradation model $d$ and the diffusion step $t$ are important factors to the final results. In our experiments, we adopt the degradation model proposed by Wang \etal \cite{wang2021realesrgan} and randomly sample $t$ from $[0, 500]$ and $[0, 250]$ for BFR and BISR, respectively. 

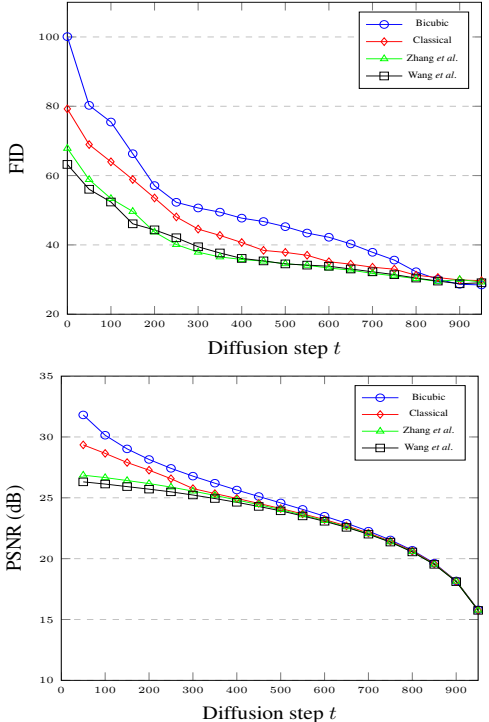
\begin{figure}[t!]
\centering
\begin{minipage}{0.46\textwidth}
\resizebox{0.8\textwidth}{0.6\textwidth}{\begin{tikzpicture}
   \begin{axis}[
       xlabel={Diffusion step $t$},
       ylabel={FID},
       xmin=0, xmax=950,
       ymin=20, ymax=110,
       xtick={0,100,200,300,400,500,600,700,800,900},
       ytick={20,40,60,80,100},
       legend pos=north east,
       ymajorgrids=true,
       grid style=dashed,
       ticklabel style={font=\tiny},
       legend style={font=\tiny}
   ]

   \addplot[
       color=blue,
       mark=o,
       ]
       coordinates {
       (0,100.09)(50,80.26)(100,75.45)(150,66.31)(200,57.11)(250,52.29)(300,50.66)(350,49.45)(400,47.72)(450,46.74)(500,45.27)(550,43.42)(600,42.21)(650,40.29)(700,37.87)(750,35.62)(800,32.25)(850,30.06)(900,28.64)(950,28.45)
       };
       \addlegendentry{Bicubic}
   \addplot[
       color=red,
       mark=diamond,
       ]
       coordinates {
       (0,79.24)(50,68.95)(100,63.98)(150,58.87)(200,53.54)(250,48.07)(300,44.56)(350,42.74)(400,40.69)(450,38.42)(500,37.84)(550,37.01)(600,35.15)(650,34.44)(700,33.57)(750,32.99)(800,31.25)(850,30.59)(900,29.88)(950,29.70)
       };
       \addlegendentry{Classical}
   \addplot[
       color=green,
       mark=triangle,
       ]
       coordinates {
       (0,67.86)(50,58.85)(100,53.36)(150,49.65)(200,43.71)(250,40.11)(300,37.90)(350,36.53)(400,35.80)(450,35.32)(500,34.56)(550,34.12)(600,33.33)(650,32.73)(700,31.78)(750,31.07)(800,30.27)(850,29.47)(900,30.02)(950,29.28)
       };
       \addlegendentry{Zhang \etal}
   \addplot[
       color=black,
       mark=square,
       ]
       coordinates {
       (0,63.22)(50,56.05)(100,52.38)(150,46.09)(200,44.34)(250,42.06)(300,39.46)(350,37.67)(400,36.10)(450,35.39)(500,34.52)(550,34.17)(600,33.85)(650,33.05)(700,32.24)(750,31.43)(800,30.44)(850,29.57)(900,28.79)(950,29.08)
       };
       \addlegendentry{Wang \etal}

   \end{axis}
   \end{tikzpicture}}
\end{minipage}
\\
\begin{minipage}{0.46\textwidth}
   \resizebox{0.8\textwidth}{0.6\textwidth}{\begin{tikzpicture}
   \begin{axis}[
       xlabel={Diffusion step $t$},
       ylabel={PSNR (dB)},
       xmin=0, xmax=950,
       ymin=10, ymax=35,
       xtick={0,100,200,300,400,500,600,700,800,900},
       ytick={10,15,20,25,30,35},
       legend pos=north east,
       ymajorgrids=true,
       grid style=dashed,
       ticklabel style={font=\tiny},
       legend style={font=\tiny}
   ]

   \addplot[
       color=blue,
       mark=o,
       ]
       coordinates {
       (50,31.81)(100,30.14)(150,29.02)(200,28.15)(250,27.42)(300,26.78)(350,26.20)(400,25.64)(450,25.11)(500,24.58)(550,24.05)(600,23.49)(650,22.91)(700,22.26)(750,21.54)(800,20.69)(850,19.63)(900,18.17)(950,15.80)
       };
       \addlegendentry{Bicubic}
   \addplot[
       color=red,
       mark=diamond,
       ]
       coordinates {
       (50,29.35)(100,28.65)(150,27.90)(200,27.27)(250,26.57)(300,25.76)(350,25.35)(400,24.95)(450,24.54)(500,24.12)(550,23.68)(600,23.21)(650,22.69)(700,22.11)(750,21.43)(800,20.63)(850,19.59)(900,18.16)(950,15.78)
       };
       \addlegendentry{Classical}
   \addplot[
       color=green,
       mark=triangle,
       ]
       coordinates {
       (50,26.86)(100,26.66)(150,26.42)(200,26.17)(250,25.88)(300,25.55)(350,25.20)(400,24.82)(450,24.44)(500,24.03)(550,23.59)(600,23.13)(650,22.61)(700,22.04)(750,21.37)(800,20.57)(850,19.54)(900,18.14)(950,15.76)
       };
       \addlegendentry{Zhang \etal}
   \addplot[
       color=black,
       mark=square,
       ]
       coordinates {
       (50,26.31)(100,26.13)(150,25.92)(200,25.71)(250,25.49)(300,25.23)(350,24.94)(400,24.63)(450,24.29)(500,23.93)(550,23.51)(600,23.07)(650,22.57)(700,22.01)(750,21.36)(800,20.55)(850,19.53)(900,18.11)(950,15.75)
       };
       \addlegendentry{Wang \etal}

   \end{axis}
   \end{tikzpicture}}
\end{minipage}
\vspace*{-4mm}
\caption{\textbf{Top:} FID versus diffusion step $t$. \textbf{Bottom:} PSNR versus diffusion step $t$.}
\label{fig:ablation}
\vspace*{-4mm}
\end{figure}

\begin{figure}[t!]
\centering
\includegraphics[width=0.48\textwidth]{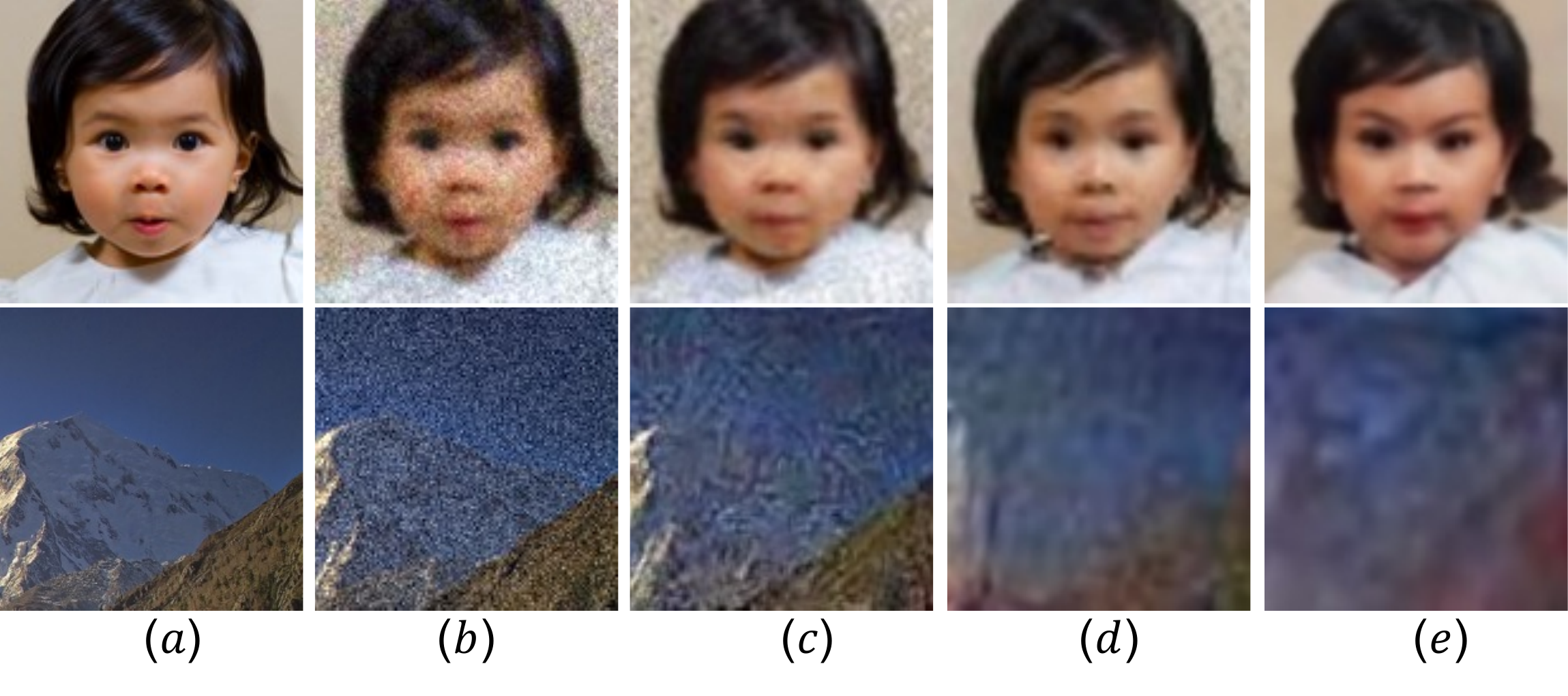}
\vspace*{-8mm}
\caption{(a) HQ image; (b) initial LQ image; (c)-(e) synthesized LQ images by our method when $t=\{250, 500, 750\}$.}
\label{fig:t}
\vspace*{-5mm}
\end{figure}

\section{Experiments}
\subsection{Experimental Setup}
To comprehensively evaluate the effectiveness of the proposed HQ-LQ training pair synthesis method, we perform experiments on both synthetic data (for quantitative evaluation) and real-world data (for qualitative evaluation). For each of the BFR and BISR tasks, we employ several open-sourced models and re-train them on our training pairs, and then apply them to the synthetic test data for objective evaluation. On the real-world test data, we invite seventeen human subjects to perform user-study and compare the visual quality of the restored HQ images.    

\textbf{Test data.}
For the task of BFR, we use the first $5,000$ HQ face images in the CelebA-HQ dataset \cite{karras2018pggan} to simulate LQ images, while for the task of BISR, we use the $100$ HQ images in the DIV2K validation dataset  \cite{timofte2017sr} to simulate LQ images. 
Apart from the synthetic test data, we also select $100$ real-world LQ face images and $100$ LQ natural images (excluded from DDPM pre-training) from the DID to qualitatively evaluate the performance of different BFR and BISR models in the wild.

\textbf{Evaluation metrics.} As for the quantitative evaluation, the Fr\'echet Inception Distances (FID) \cite{heusel2017fid}, the Learned Perceptual Image Patch Similarity (LPIPS) \cite{zhang2018lpips}, the Peak Signal-to-Noise Ratio (PSNR) and the Structural Similarity Index Measure (SSIM) \cite{wang2004ssim} indices are used to measure the distance between the model output and ground-truth. 

\textbf{Training details.}
When training a DDPM, we adopt the Adam optimizer \cite{kingma2015adam} with a batch size of $16$. The learning rate is fixed as $8\times 10^{-5}$. In particular, we adopt the exponential moving average (EMA) method with a decay coefficient of $0.995$ during optimization to ensure the training stability. The model is updated for $700K$ iterations.

In regarding to the training of downstreaming task (\ie, BFR and BISR) models, we simply keep the training configurations in their original papers \cite{chen2021psfrgan,yang2020hifacegan,wang2021gfpgan,yang2021gpen,wang2021realesrgan} unchanged, except when we train these models by using the HQ-LQ training pairs synthesized by our method.

\begin{table*}
\centering
\caption{The FID, LPIPS, PSNR and SSIM indices of original BFR models and their re-trained counterparts (denoted by ``+") on the synthesized test data.}
\vspace*{-3mm}
    \resizebox{0.98\textwidth}{!}{\begin{tabular}{l|c c|c c|c c|c c}
      Metric & PSFRGAN \cite{chen2021psfrgan} & PSFRGAN+ & HiFaceGAN \cite{yang2020hifacegan} & HiFaceGAN+ & GFPGAN \cite{wang2021gfpgan} & GFPGAN+ & GPEN \cite{yang2021gpen} & GPEN+ \\
      \hline \hline
      {FID$\downarrow$} & \textbf{13.7655} & 14.9023 & 17.6288 & \textbf{12.4294} & 10.1204 & \textbf{10.1057} & 8.0718 & \textbf{6.8306} \\
      \hline
      {LPIPS$\downarrow$} & 0.2720 & \textbf{0.2717} & 0.2783 & \textbf{0.2588} & 0.2252 & \textbf{0.2161} & 0.2618 & \textbf{0.2299} \\
      \hline
      {PSNR$\uparrow$} & 25.5152 & \textbf{25.7500} & \textbf{26.3227} & 25.1443 & 25.1169 & \textbf{25.7269} & 24.9363 & \textbf{26.2205} \\
      \hline
      {SSIM$\uparrow$} & 0.6496 & \textbf{0.6633} & 0.6505 & \textbf{0.6802} & 0.6739 & \textbf{0.6813} & 0.6276 & \textbf{0.6775} \\
   \end{tabular}}
\label{tab:bfr}
\vspace*{-2mm}
\end{table*}

\begin{figure*}[t!]
\centering
\includegraphics[width=0.98\textwidth]{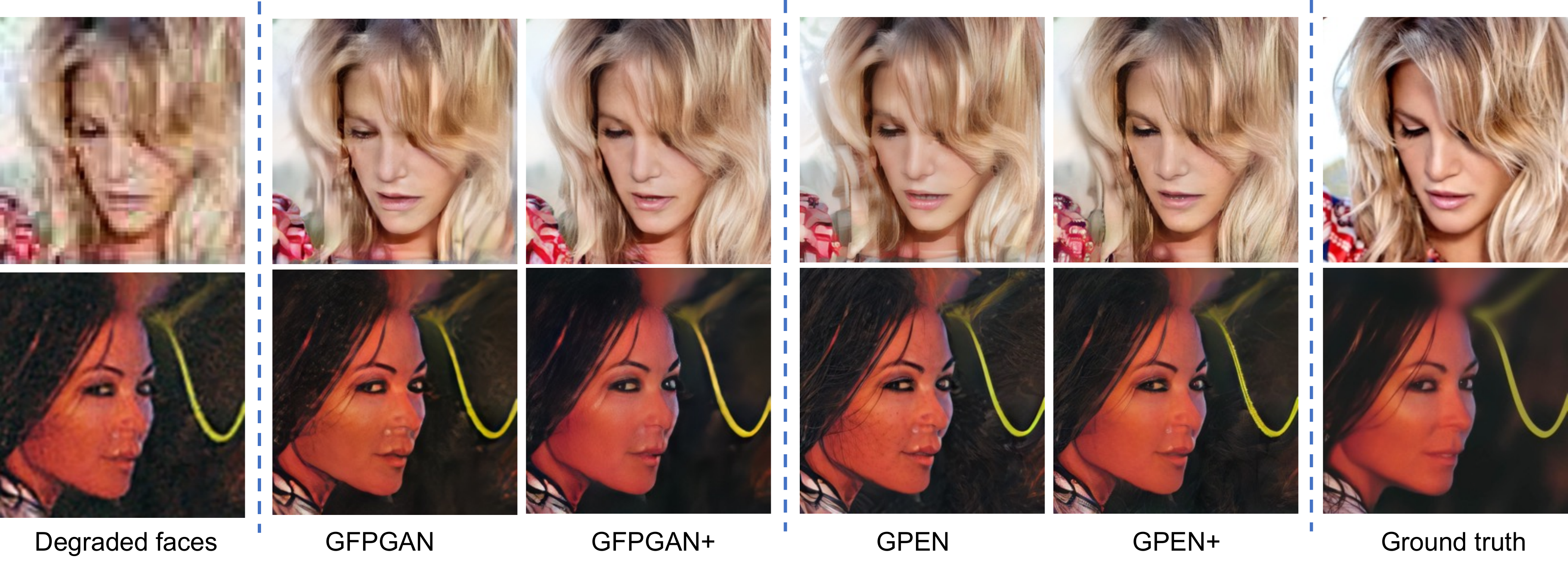}
\vspace{-3mm}
\caption{Blind face restoration results of original BFR models and their re-trained counterparts (denoted by ``+") by our synthesized training pairs. 
}
\label{fig:bfr}
\vspace*{-3mm}
\end{figure*}

\subsection{Selections of Initial Handcrafted Degradation and Diffusion Step}
In order to find out the effects of handcrafted degradation model $d$ and diffusion step $t$ on the synthesized HQ-LQ pairs, we conduct ablation studies by applying four representative handcrafted degradation models, \ie, bicubic downsampling (denoted by Bicubic), the classical degradation model (denoted by Classical, see \cite{lai2017lapsrn,yang2021gpen}), the one proposed by Zhang \etal \cite{zhang2021bsrgan} and the one proposed by Wang \etal \cite{wang2021realesrgan}, to our method. 

We use the first $5,000$ face images in FFHQ \cite{karras2019stylegan2} as the HQ inputs, and generate four groups of LQ counterparts by our method with the four handcrafted degradation models. We calculate the FID \cite{heusel2017fid} between each group of generated LQ images and the real degraded faces in our collected DID to evaluate the distribution distances.

\textbf{Handcrafted Degradation Model.}
The curves of FID versus diffusion step $t$ are plotted on the top of Fig. \ref{fig:ablation}. We see that when $t=0$ (\ie, without applying the pre-trained DDPM), Bicubic has the worst FID, while Zhang \etal \cite{zhang2021bsrgan} and Wang \etal \cite{wang2021realesrgan} are among the best. This indicates that the degradation models proposed by Zhang \emph{et al.} \cite{zhang2021bsrgan} and Wang \emph{et al.} \cite{wang2021realesrgan} can better simulate real-world degraded data than bicubic downsampling and the classical degradation model. It is worth mentioning that all the four models can produce realistic LQ images (\ie, low FID value) with a large enough step number $t$; however, a good degradation model should be able to generate LQ images that cover a large scope of LQ space. The models developed by Zhang \etal \cite{zhang2021bsrgan} and Wang \etal \cite{wang2021realesrgan} demonstrate their superiority to bicubic downsampling and the classical degradation model. We choose the model proposed by Wang \etal \cite{wang2021realesrgan} due to its efficient implementation. 

\textbf{Diffusion Step.} A qualified HQ-LQ image pair should share the same content and structure in general. We calculate the PSNR of generated HQ-LQ pairs and plot the curve on the bottom of Fig. \ref{fig:ablation}.
One can see that all curves converge with the increase of diffusion step $t$. This is because the distributions of LQ images and real-world degraded images are getting closer during the diffusion process, and they approach to the same isotropic Gaussian distribution. In other words, with a large enough diffusion step $t$, we can produce LQ images with lower FID values. However, a large $t$ will destroy the structure of synthesized LQs compared with their HQ counterparts. As shown in Fig.~\ref{fig:ablation}, the PSNR decreases when $t$ grows. By our experiments, when the PSNR of an HQ-LQ pair is smaller than $24$, the general structure of them will become inconsistent. Fig.~\ref{fig:t} shows some synthesized LQ images by our method with different steps $t$. One can see that the structures of synthesized LQ images become inconsistent with the HQ ones when $t>500$ / $t>250$ for face/natural images.
We therefore randomly sample $t$ from $[0, 500]$ for the training pair synthesis of BFR, and sample $t$ from $[0, 250]$ for BISR.  

\begin{table*}
\Huge
\centering
\caption{The FID, LPIPS, PSNR and SSIM indices of different BISR models on the synthesized test data. RRDB+ refer to the RRDB-based model trained on our training pairs.}
\vspace*{-3mm}
    \Huge{
    \resizebox{0.98\textwidth}{!}{\begin{tabular}{l|c c c c c|c c c|c}
      Metric & ESRGAN-FS \cite{fritsche2019frequency} & DASR \cite{wei2021dasr} & Wang \etal \cite{wang2021unsupervisedsr} & PDM-SRGAN \cite{luo2022pdmsr} & F2N-ESRGAN \cite{li2022face2natual} & RealSR \cite{ji2019realsr} & BSRGAN \cite{zhang2021bsrgan} & Real-ESRGAN \cite{wang2021realesrgan} & RRDB+ \\
      \hline\hline
      {FID$\downarrow$} & 106.1637 & 100.9717 & 110.1927 & 102.8514 & 100.3145 & 115.2377 & 93.0567 & 91.9363 & \textbf{89.2679} \\
      \hline
      {LPIPS$\downarrow$} & 0.5168 & 0.4789 & 0.5232 & 0.4621 & 0.4349 & 0.6655 & 0.4287 & \textbf{0.4223} & 0.4248 \\
      \hline
      {PSNR$\uparrow$} & 20.9974 & 21.1608 & 20.2800 & 21.0749 & 21.132 & 21.0939 & 21.8632 & 21.3071 & \textbf{22.1809}  \\
      \hline
      {SSIM$\uparrow$} & 0.5194 & 0.5099 & 0.4927 & 0.5226 & 0.5417 & 0.4629 & 0.5439 & 0.5389 & \textbf{0.5599}  \\
   \end{tabular}}}
\label{tab:bisr}
\vspace*{-2mm}
\end{table*}

\begin{figure*}[t!]
\centering
\includegraphics[width=0.98\textwidth]{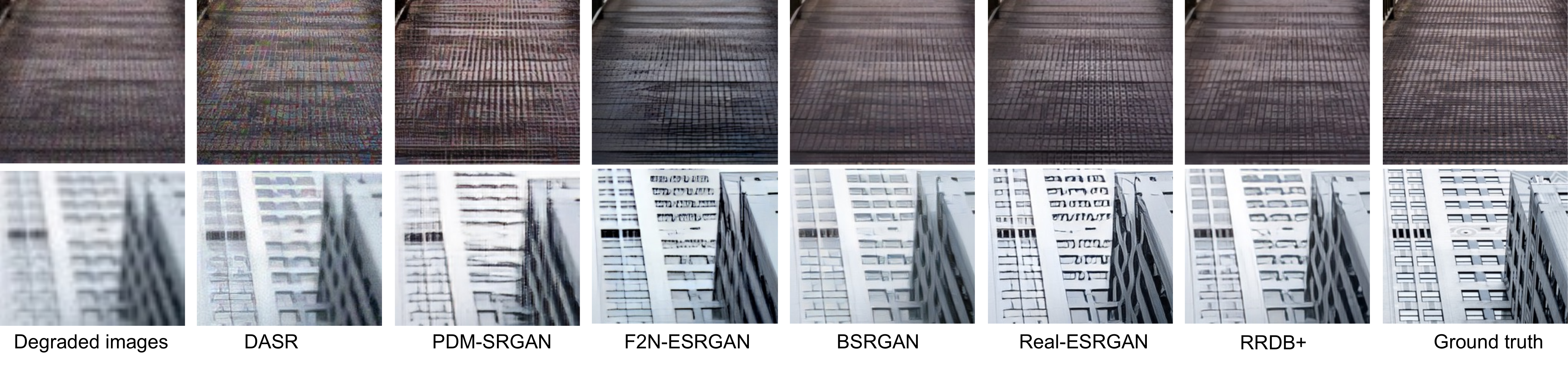}
\vspace{-2mm}
\caption{Blind image super-resolution results on synthesized degraded images. All models use the RRDB \cite{wang2018esrgan} backbone. The performance differences are mainly caused by different training pairs. 
}
\label{fig:bisr}
\vspace*{-3mm}
\end{figure*}

\begin{figure}[t!]
\centering
\includegraphics[width=0.48\textwidth]{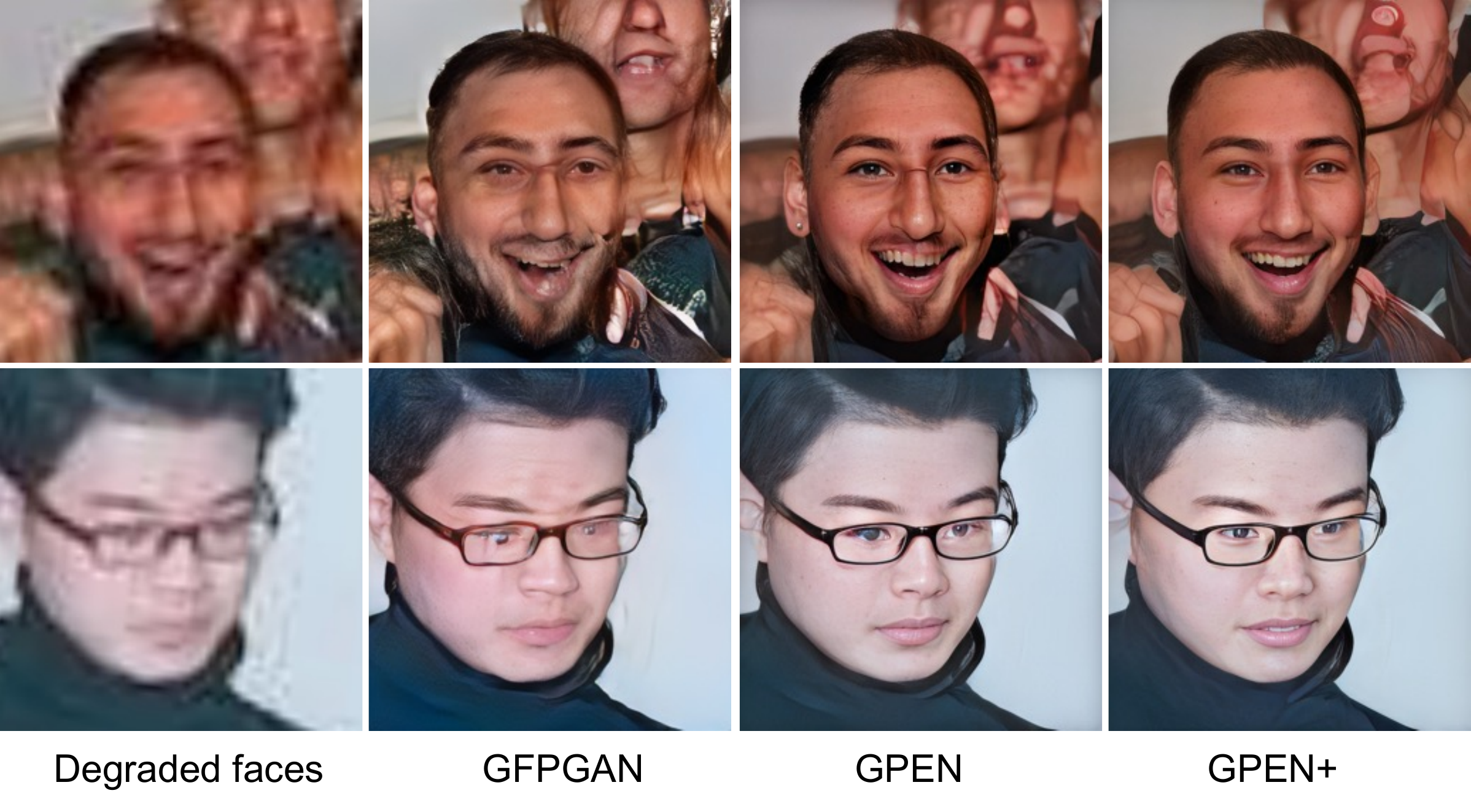}
\vspace{-6mm}
\caption{Blind face restoration results on real degraded faces in the wild. 
}
\label{fig:bfr-real}
\vspace*{-3mm}
\end{figure}

\begin{figure}[t!]
\centering
\includegraphics[width=0.48\textwidth]{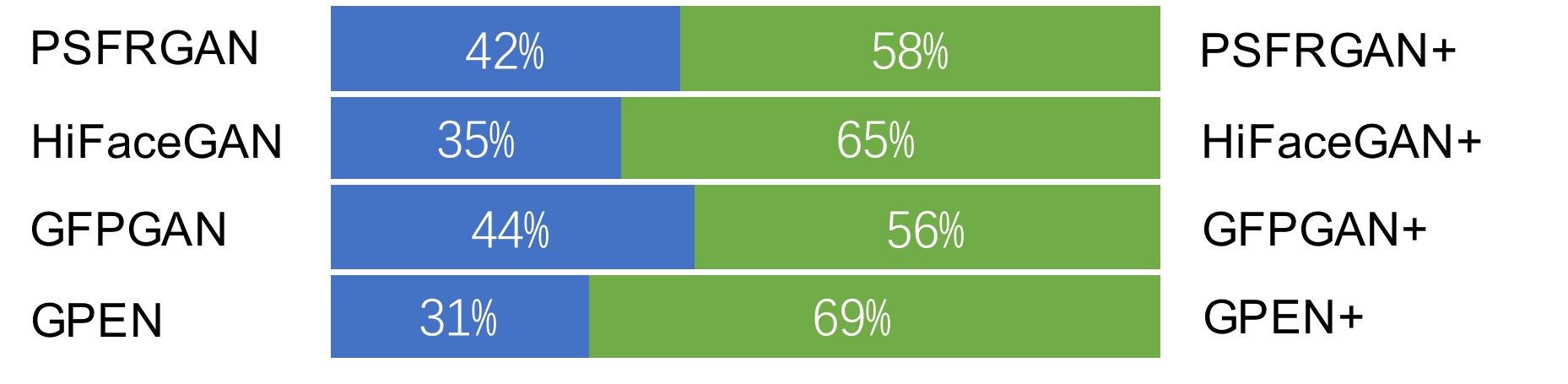}
\vspace*{-6mm}
\caption{User preferences of original BFR models versus the re-trained models on our synthesized training pairs.}
\label{fig:ustudy}
\vspace*{-3mm}
\end{figure}

\begin{figure*}[t!]
\centering
\includegraphics[width=0.98\textwidth]{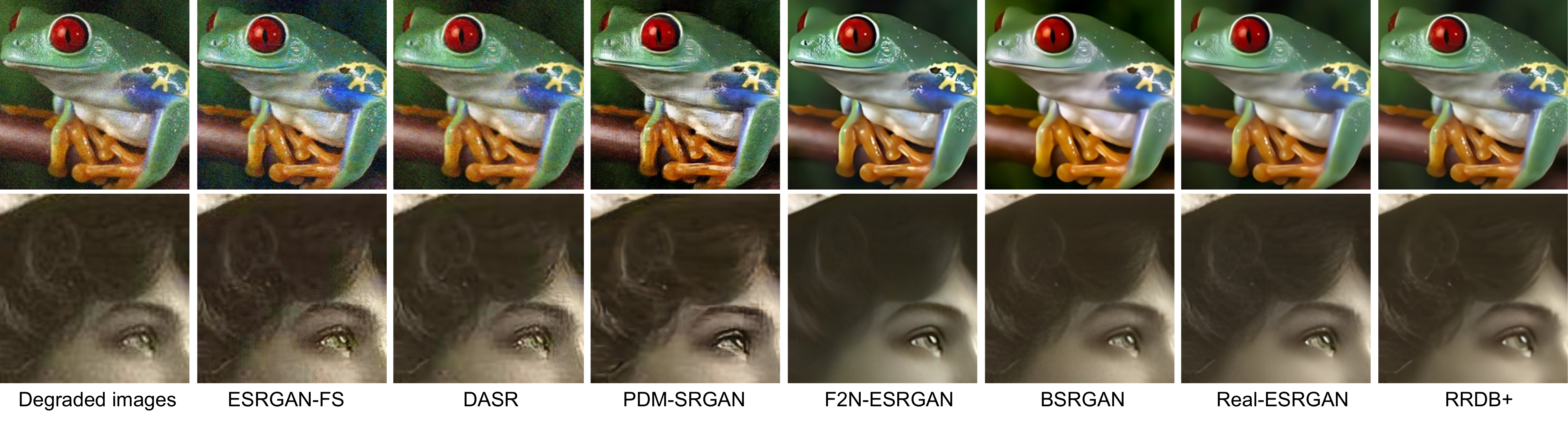}
\vspace{-4mm}
\caption{Blind image super-resolution results on degraded images in the wild. 
}
\label{fig:bisr-real}
\vspace*{-3mm}
\end{figure*}

\begin{figure}[t!]
\centering
\includegraphics[width=0.48\textwidth]{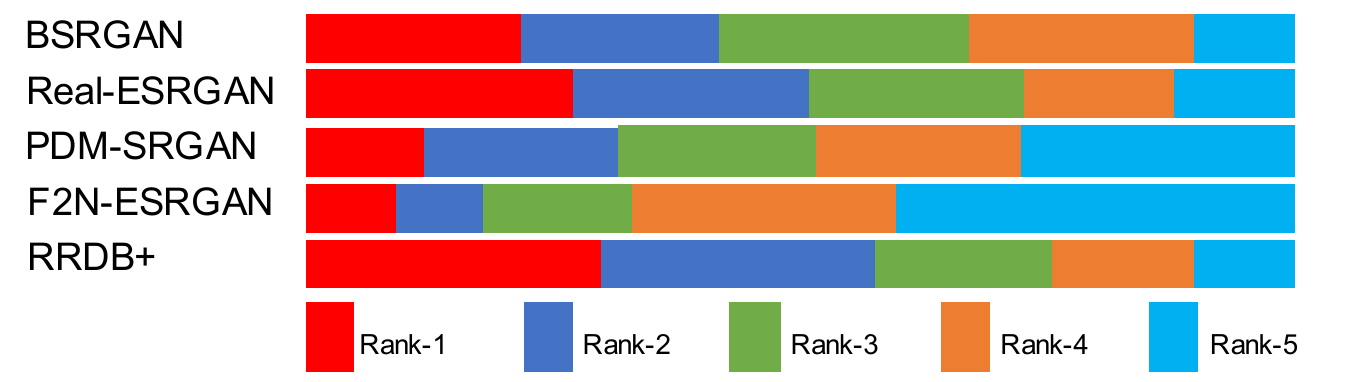}
\vspace*{-6mm}
\caption{User study results of different BISR methods.}
\label{fig:ustudy2}
\vspace*{-3mm}
\end{figure}

\subsection{Experiments on Synthetic Test Images}
\label{sec:experiment_syn}
\textbf{Blind Face Restoration.}
We validate the effectiveness of our approach by using four representative BFR models, \ie, PSFRGAN \cite{chen2021psfrgan}, HiFaceGAN \cite{yang2020hifacegan}, GFPGAN \cite{wang2021gfpgan} and GPEN \cite{yang2021gpen}, whose training codes are publicly available so that we can re-train them by using the HQ-LQ image pairs synthesized by our method with their original training settings. We denote by PSFRGAN+, HiFaceGAN+, GFPGAN+ and GPEN+ the re-trained models on our training data. By applying those original and re-trained models on the synthetic LQ face images, we can calculate the FID, LPIPS, PSNR and SSIM indices by comparing the reconstructed HQ images with the ground-truth HQ images. The results are listed in Table~\ref{tab:bfr}. One can see that for each BFR method, the model trained on our synthesized training pairs can achieve better FID/LPIPS/PSNR/SSIM indices than its original counterpart in most cases. 

Fig.~\ref{fig:bfr} compares the BFR results on some degraded face images by original BFR models and their re-trained counterparts. It can be seen that GFPGAN \cite{wang2021gfpgan} and GPEN \cite{yang2021gpen} can produce reasonable face reconstructions while failing to tackle faces that are severely compressed or with large viewpoint variation. After re-training GFPGAN and GPEN with HQ-LQ training pairs synthesized by the proposed method, GFPGAN+ and GPEN+ succeed in restoring clearer faces with more vivid details in comparison with their original counterparts. 
More visual comparison results can be found in Sec.~\ref{sec:supp-bfr}. 

\textbf{Blind Image Super-Resolution.}
BISR aims to reconstruct HQ images with perceptually realistic details from the input LQ image with unknown degradation. We employ two categories of state-of-the-art algorithms in this experiment. The first category is GAN-based methods trained on unpaired real data, including ESRGAN-FS \cite{fritsche2019frequency}, DASR \cite{wei2021dasr}, Wang \etal \cite{wang2021unsupervisedsr}, PDM-SRGAN \cite{luo2022pdmsr}, F2N-ESRGAN \cite{li2022face2natual}. All methods are re-trained on our DID. The second category is models trained with handcrafted HQ-LQ image pairs, including RealSR \cite{ji2019realsr}, BSRGAN \cite{zhang2021bsrgan}, Real-ESRGAN \cite{wang2021realesrgan}. All these three methods use RRDB \cite{wang2018esrgan} as the backbone with similar losses. The key difference among them lies in the training data. Therefore, we re-train a RRDB-based model \cite{wang2018esrgan} by using our synthesized training pairs with the training settings of Real-ESRGAN \cite{wang2021realesrgan}, resulting in the RRDB+ model.  The quantitative evaluation results on the test data are presented in Table~\ref{tab:bisr}. One can see that RRDB+ outperforms the competing methods in most of the metrics. In particular, compared with the original Real-ESRGAN model, our re-trained  RRDB+ achieves similar LPIPS index but significantly better FID, PSNR and SSIM indices. 

Fig.~\ref{fig:bisr} presents the visual comparison of competing BISR methods on several synthesized LQ natural images. One can see that our method can better reconstruct realistic details and preserve the general structures, demonstrating the effectiveness of our HQ-LQ training pair synthesis methods. Comparing to Real-ESRGAN \cite{wang2021realesrgan}, RRDB+ has clear advantages in inhibiting artifacts. 

\subsection{Experiments on Images in the Wild}
Since our method targets at synthesizing realistic image restoration training pairs, it is necessary to conduct experiments on images in the wild. We select 100 face and 100 natural images from DID for evaluation. 

\textbf{Blind Face Restoration.}
We apply the same four BFR methods as in Sec.~\ref{sec:experiment_syn} to the 100 real-world LQ face images. Fig.~\ref{fig:bfr-real} shows the BFR results on two images. Due to the limit of space, we only show the results of GFPGAN \cite{wang2021gfpgan}, GPEN \cite{yang2021gpen} and GPEN+ here. More results can be found in Sec.~\ref{sec:supp-bfr-real}. 
As can be seen, GPEN+ can better handle real-world degradations and produce more realistic results than GPEN and GFPGAN, mainly due to the more realistic image training pairs.

To better evaluate the advantage of the proposed method, we conduct a user study as subjective assessment. For each of the four groups of BFR methods, \ie, PSFRGAN vs. PSFRGAN+, HiFaceGAN vs. HiFaceGAN+, GFPGAN vs. GFPGAN+, GPEN vs. GPEN+, the BFR results are presented in pairs to $17$ volunteers in a random order. The volunteers are asked to pick the better BFR image from each pair according to their perceptual quality. The statistics are presented in Fig.~\ref{fig:ustudy}. One can see that in each group, the model trained with our synthesized data is more preferred.

\textbf{Blind Image Super-Resolution.}
Fig.~\ref{fig:bisr-real} shows the BISR results of ESRGAN-FS \cite{fritsche2019frequency}, DASR \cite{wei2021dasr}, BSRGAN \cite{zhang2021bsrgan}, Real-ESRGAN \cite{wang2021realesrgan}, PDM-SRGAN \cite{luo2022pdmsr}, F2N-ESRGAN \cite{li2022face2natual}, and our RRDB+. It can be seen that the competing methods yield dirty outputs, \eg, the second and third columns of Fig.~\ref{fig:bisr-real}, or tend to produce over-smoothed results, \eg, the frog in the first row of Fig.~\ref{fig:bisr-real}, or fail to reconstruct photo-realistic details, \eg, the eye in the second row of Fig.~\ref{fig:bisr-real}. With the help of the realistic HQ-LQ training pairs generated by our method, RRDB+ is capable of better handling complex degradations in the wild. More visual results can be found in Sec.~\ref{sec:supp-bisr-real}. 

Since the commonly used quantitative metrics do not correlate well with human visual perception to image quality, we conduct a user study as subjective assessment. The BISR results of BSRGAN \cite{zhang2021bsrgan}, Real-ESRGAN \cite{wang2021realesrgan}, PDM-SRGAN \cite{luo2022pdmsr}, F2N-ESRGAN \cite{li2022face2natual} and RRDB+ on $100$ natural images from DID are presented to $17$ volunteers in random order. The volunteers are asked to rank the five BISR outputs of each input image according to their perceptual quality. As presented in Fig.~\ref{fig:ustudy2}, our method RRDB+ receives the most rank-1 votes and the least rank-5 votes.


\section{Conclusion and Discussion}
We proposed, for the first time to our best knowledge, to train a DDPM to synthesize realistic image restoration training pairs. With a collected LQ image dataset, a DDPM was first trained to approximate its distribution. The pre-trained DDPM was then employed to convert the initially degraded image from its HQ counterpart into the desired LQ image, which fell into the distribution of real-world LQ images. With the synthesized realistic HQ-LQ image pairs, we re-trained the many state-of-the-art BFR and BISR models, and the re-trained models demonstrated much better realistic image restoration performance than their original counterparts both quantitatively and qualitatively.

It should be noted that in our experiments, we collected a large scale degraded face dataset and a natural image dataset, respectively, as the target distributions to train the DDPM and synthesize HQ-LQ training pairs. In practice, the users can build their own LQ image dataset according to their needs, train the corresponding DDPM models and synthesize training pairs for different image restoration tasks. 

{\small
\bibliographystyle{ieee_fullname}
\bibliography{egpaper_final}
}

\begin{alphasection}
\section*{Appendices}

\subsection{More BFR Results on Synthetic Faces}
\label{sec:supp-bfr}
This section shows more visual results of competing BFR methods on synthetic face images. As in the main paper, we compare the BFR results on degraded face images by the original BFR models, \ie, GFPGAN \cite{wang2021gfpgan} and GPEN \cite{yang2021gpen}, and their re-trained counterparts on our training pairs, \ie, GFPGAN+ and GPEN+. The visual comparisons are presented in Fig.~\ref{fig:bfr}. One can see that GFPGAN+ and GPEN+ generate superior results with finer details to their original counterparts. 

\subsection{More BFR Results on Faces in the Wild}
\label{sec:supp-bfr-real}
This section shows more visual results of competing BFR methods on face images in the wild. As in the main paper, we compare our method with PSFRGAN \cite{chen2021psfrgan}, HiFaceGAN \cite{yang2020hifacegan}, GFPGAN \cite{wang2021gfpgan} and GPEN \cite{yang2021gpen}. The visual comparisons are presented in Fig.~\ref{fig:bfr-real}. It can be seen that our method can produce more realistic results.

\subsection{More BISR Results on Images in the Wild}
\label{sec:supp-bisr-real}
In this section, we show more visual comparison between our method with state-of-the-art BISR methods, including DASR \cite{wei2021dasr}, PDM-SRGAN \cite{luo2022pdmsr}, BSRGAN \cite{zhang2021bsrgan} and Real-ESRGAN \cite{wang2021realesrgan}, in Fig.~\ref{fig:bisr-real}, from which we can see the superior performance of our method on BISR tasks. 

\subsection{Comparison between GAN-based methods and our diffusion method on LQ image synthesis}
In order to demonstrate the superior performance of our proposed diffusion-based method in synthesizing LQ images to existing GAN-based methods \cite{fritsche2019frequency,wei2021dasr,luo2022pdmsr,li2022face2natual}, we compute the FID values to evaluate the distribution distances between synthesized LQs and real LQs in DID. The quantitative results are presented in Table~\ref{tab:gap}. It can be seen that our diffusion-based method outperforms existing GAN-based methods significantly. This is because GAN-based methods learn a direct mapping from HQs to LQs, which is hard to train due to the large distribution gap and unstable adversarial training, while our method employs an initial LQ image as the starting point, and use a pre-trained diffusion model to stably generate the final LQ image. During the reverse process of DDPM, the images are naturally constrained in the LQ space. In contrast, GAN-based methods only constrain the final outputs, and hence they are less stable in LQ image synthesis.
\end{alphasection}

\begin{table*}[t!]
\centering
\caption{Comparison (FID) of different methods in synthesizing LQ images.}
\vspace*{-3mm}
   \resizebox{0.88\textwidth}{!}{\begin{tabular}{c|c|c|c|c|c}
      Method & ESRGAN-FS \cite{fritsche2019frequency} & DASR \cite{wei2021dasr} & PDM-SRGAN \cite{luo2022pdmsr} & F2N-ESRGAN \cite{li2022face2natual} & Ours \\
      \hline
      {FID$\downarrow$} & 79.5057 & 77.7190 & 68.8146 & 70.4900 & \textbf{49.2831} \\
   \end{tabular}}
\label{tab:gap}
\vspace*{-2mm}
\end{table*}

\begin{figure*}[t!]
\centering
\includegraphics[width=0.94\textwidth]{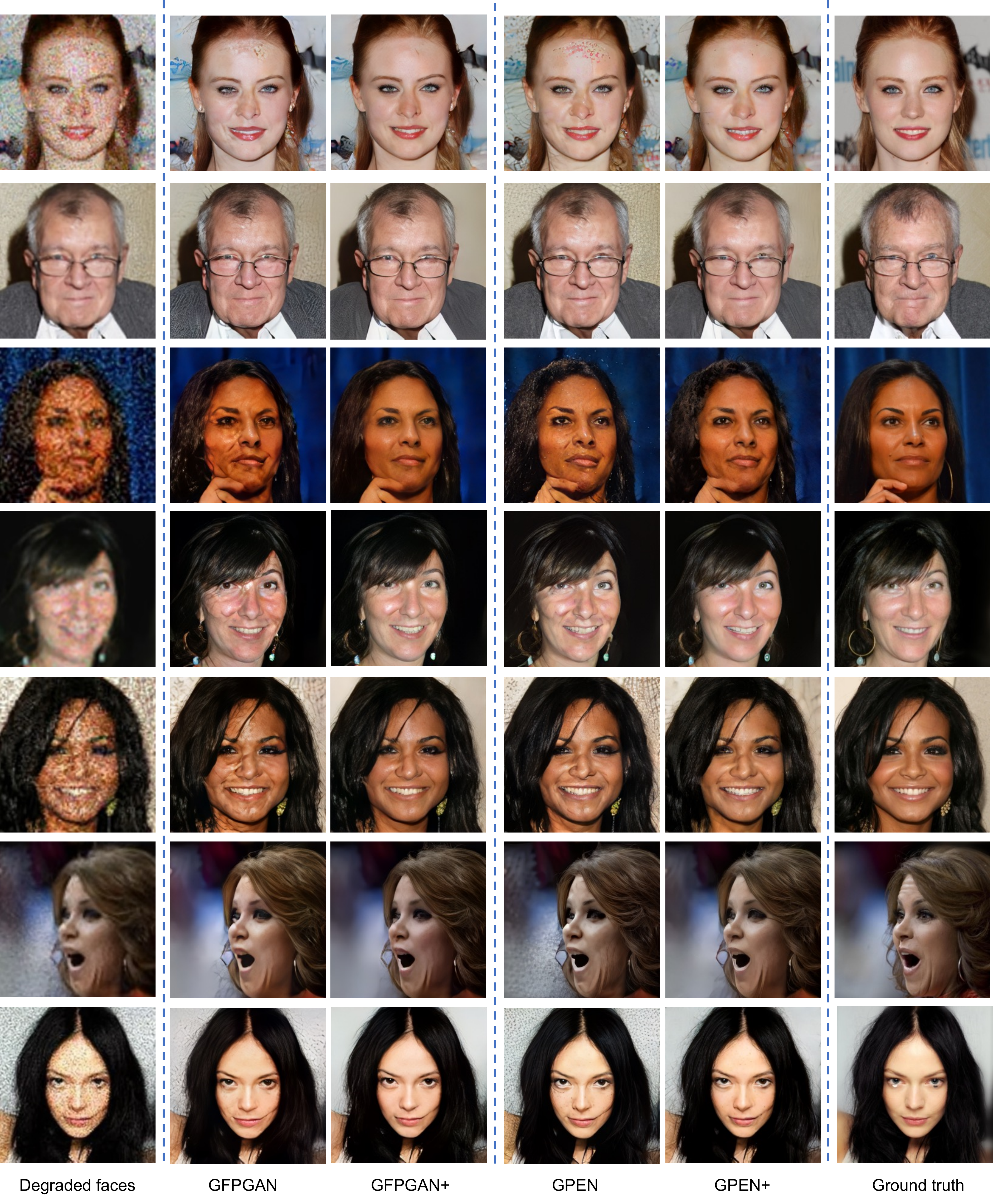}
\vspace*{-4mm}
\caption{Blind face restoration results of original BFR models and their re-trained counterparts (denoted by ``+") by our synthesized training pairs.} 
\label{fig:bfr}
\end{figure*}

\begin{figure*}[t!]
\centering
\includegraphics[width=0.98\textwidth]{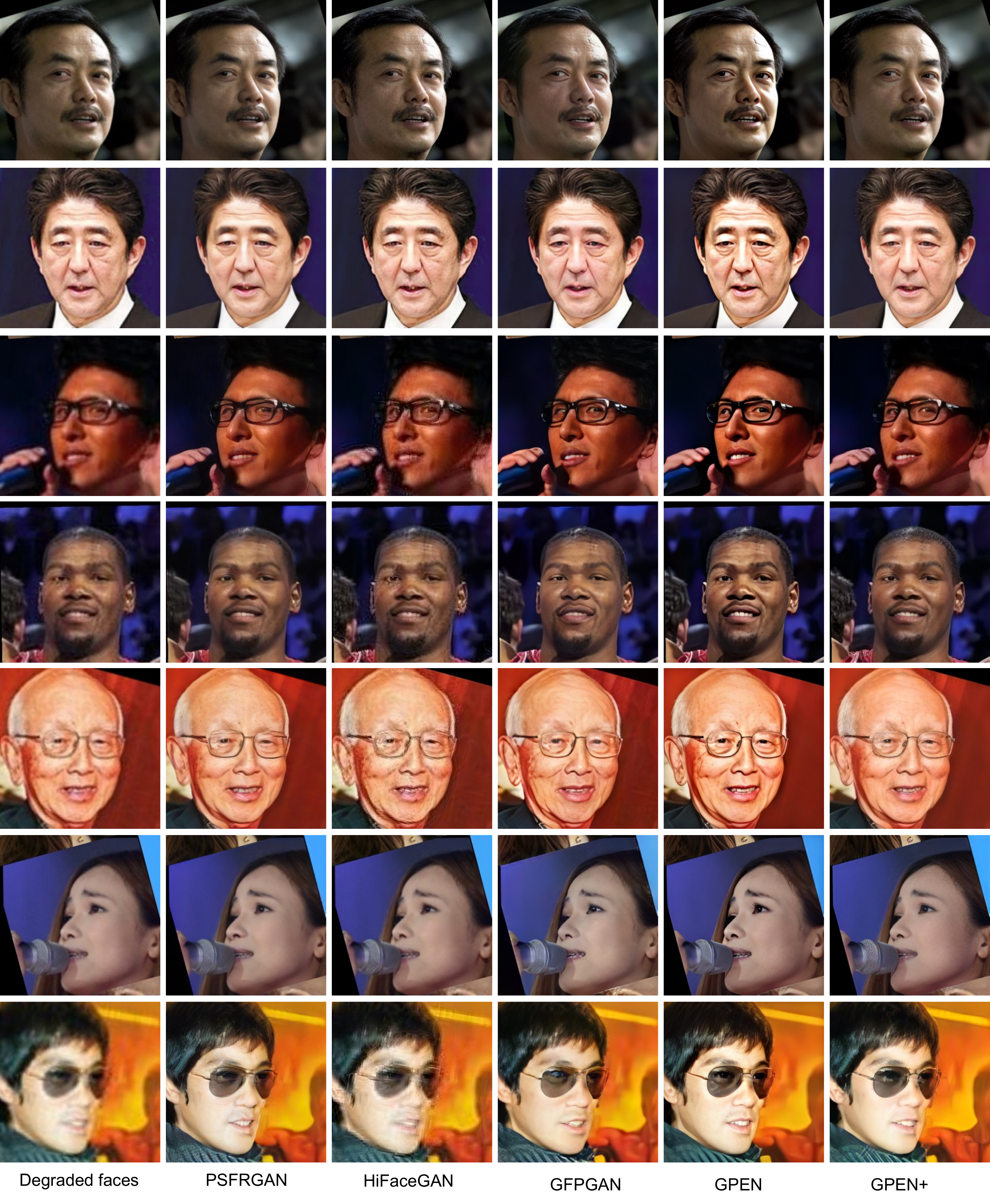}
\caption{Blind image super-resolution results on real degraded faces in the wild. }
\label{fig:bfr-real}
\end{figure*}

\begin{figure*}[t!]
\centering
\includegraphics[width=0.98\textwidth]{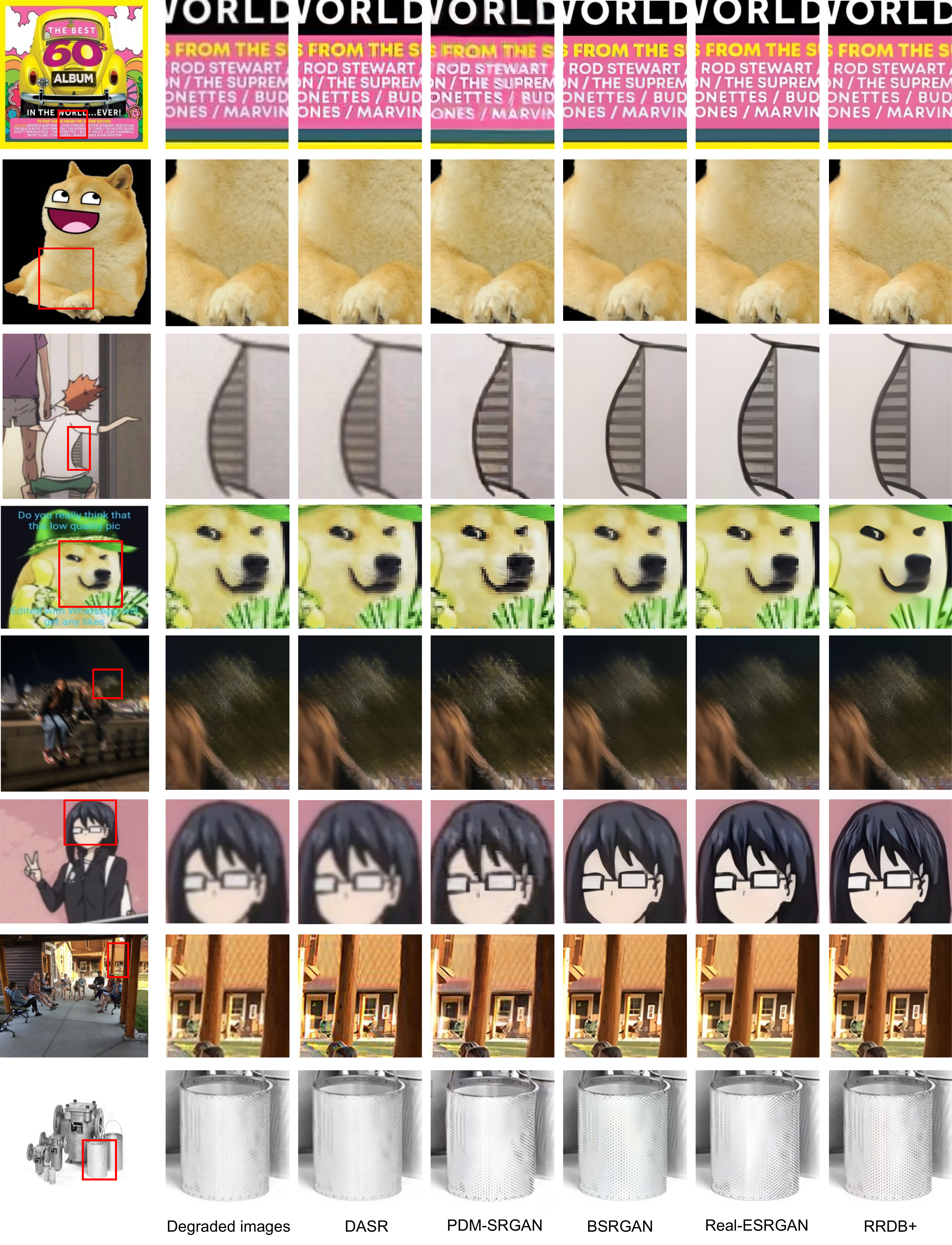}
\vspace*{-4mm}
\caption{Blind image super-resolution results on degraded images in the wild.}
\label{fig:bisr-real}
\end{figure*}

\end{document}